\documentclass[journal]{IEEEtran}
\usepackage{graphicx}   
\usepackage{amsmath}
\usepackage{multirow,rotating}
\usepackage{color}
\usepackage{soul}
\usepackage{stfloats}
\usepackage{flushend}

\begin{document}
\title{Disjunctive Normal Networks}
\author{Mehdi~Sajjadi,~\IEEEmembership{}
        Mojtaba~Seyedhosseini~\IEEEmembership{}
                and~Tolga~Tasdizen,~\IEEEmembership{Senior Member,~IEEE,}
\thanks{The authors are with the Department of Electrical and Computer Engineering, University of Utah, Salt Lake City, UT, 84112 USA e-mail: mehdi@sci.utah.edu}}

\setstcolor{red}
\maketitle
\setcounter{page}{1}
\begin{abstract}
Artificial neural networks are powerful pattern classifiers; however, they have been surpassed in accuracy by methods such as support vector machines and random forests that are also easier to use and faster to train. Backpropagation, which is used to train artificial neural networks, suffers from the herd effect problem which leads to long training times and limit classification accuracy. We use the disjunctive normal form and approximate the boolean conjunction operations with products to construct a novel network architecture. The proposed model can be trained by minimizing an error function and it allows an effective and intuitive initialization which solves the herd-effect problem associated with backpropagation. This leads to state-of-the art classification accuracy and fast training times. In addition, our model can be jointly optimized with convolutional features in an unified structure leading to state-of-the-art results on computer vision problems with fast convergence rates. A GPU implementation of LDNN with optional convolutional features is also available
\end{abstract}

%

%
\IEEEpeerreviewmaketitle

\section{Introduction}
\label{sec:intro}
%
%
%
%
\IEEEPARstart{A}{n} artificial neural network (ANN) consisting of one hidden layer of squashing functions is an universal approximator for continuous functions defined on the unit hypercube~\cite{Hornik1989,Cybenko1989}. However, until the introduction of the backpropagation algorithm~\cite{Rumelhart1986}, training such multilayer perceptron (MLP) networks was not possible in practice. The backpropagation algorithm propelled MLPs to be the method of choice for many classification and regression applications. However, eventually MLPs were replaced by more recent techniques such as support vector machines (SVM)~\cite{Cortes1995} and random forests (RF)~\cite{Breiman2001}. In addition to being surpassed in accuracy by these modern techniques, an important drawback of MLPs has been the high computational cost of training emphasized by growing data set sizes and dimensionality. An underlying reason for the limited accuracy and high computational cost of training is the herd-effect problem~\cite{Fahlman90thecascade-correlation}. During backpropagation each hidden unit tries to evolve into a useful feature detector from a random initialization; however, this task is complicated by the fact that all units are changing at the same time without any direct communication between them. Consequently, hidden units can not effectively subdivide the necessary computational tasks among themselves leading to a complex dance which can take a long time to settle down.  


In this paper, we introduce a new network architecture that overcomes the difficulties associated with MLPs and backpropagation for supervised learning. Our network consists of one adaptive layer of feature detectors implemented by logistic sigmoid functions followed by two fixed layers of logical units that compute conjunctions and disjunctions, respectively. We call the proposed network architecture Logistic Disjunctive Normal Network (LDNN). Unlike MLPs, LDNNs allow for a simple and intuitive initialization of the network weights which avoids the herd-effect. Furthermore, due to the single adaptive layer, it allows larger step sizes in minimizing the error function.
We also propose a deep learning structure which consists of automatic convolutional feature extractors and LDNNs as efficient classifiers. The proposed structure performs automatic feature extraction and classification simultaneously and in an unified structure. Finally, we present results of experiments on LDNN for general classification and image classification using proposed deep structure. For general classification, we conducted experiments on 10 binary and 6 multi-class classification problems. LDNNs outperformed MLPs in every case both in terms of accuracy and computational speed. LDNNs produced the best accuracy in 11 out of the 16 classification problems in comparison to SVMs and RFs. For image classification, we tested our deep structure on 5 popular datasets. Our model was able to achieve state-of-the-art performance on 2 out of 5 datasets and competitive results on the rest.

\section{Related Work}
\label{sec:related}
Extensive research has been performed on variants of the backpropagation algorithm including batch vs. stochastic learning~\cite{Heskes1993,Orr1995}, squared error vs. cross-entropy~\cite{Joost1998} and optimal learning rates~\cite{Saad1995,Murata1997}. Many other practical choices including normalization of inputs, initialization of weights, stopping criteria, activation functions, target output values that will not saturate the activation functions, shuffling training examples, momentum  terms in optimization, and optimization techniques that make use of the second-order derivatives of the error are summarized in~\cite{LeCun1998}. More recently, Hinton {\em et al.} proposed a Dropout scheme for backpropagation which helps prevent co-adaptation of feature detectors~\cite{HintonDropout}. Despite the extensive effort devoted to making learning MLPs as efficient as possible, the fundamental problems outlined in Section~\ref{sec:intro} remain because they arise from the architecture of MLPs.  Contrastive divergence~\cite{Hinton2006b,Hinton2006a} can be used to pre-train networks in an unsupervised manner prior to backpropagation such that the herd-effect problem is alleviated. Contrastive divergence has been used successfully to train deep networks. The LDNN model proposed in this paper can be seen as an architectural alternative for supervised learning of ANNs.

The idea of representing classification functions in disjunctive form has been previously explored in the literature. Fuzzy min-max networks~\cite{Simpson1992,Song1993,Nandedkar2004} represent the classification function as the union of axis aligned hypercubes in the feature space. The most important drawback of this model is its limitation to axis aligned decision boundaries which can significantly increase the number of conjunctions necessary for a good approximation. We construct a significantly more efficient approximation by using an union of convex polytopes. Furthermore, fuzzy min-max neural networks employ an adhoc expansion-contraction scheme for learning, whereas we formulate learning as an energy minimization problem. Lu {\em et al.}~\cite{Lu1994} proposed a multi-sieving network that decomposes learning tasks. Lee {\em et al.}~\cite{Lee1998} proposed a disjunctive fuzzy network which is based on prototypes; however, it lacks an objective function and is based on an adhoc training procedure. Similarly, the modular network proposed by Lu and Ito~\cite{Lu1999} removes the axis aligned hypercube restriction from fuzzy min-max networks; however, their network can not be learned by minimizing a single energy function. Our LDNN model uses differentiable activation functions which makes it possible to optimize the network parameters in an unified manner by minimizing a single energy function. We show that unified training of our classifier results in very significant accuracy advantages over the modular network. Differentiable approximations of min-max functions have been used to construct fuzzy neural network that can be trained using steepest descent~\cite{Marks1992,Normura1992,Zhang1994,Zhang1996}, but these have produced results that are significantly less accurate than state-of-the-art classification techniques. A closely related approach to ours is adaptive mixtures of local experts which uses a gating network to stochastically select the output from a set of feedforward networks~\cite{Jacobs1991}. The reader is referred to ~\cite{Yuksel2012} for a survey of mixture of expert methods. The products of experts approach models complex probability distributions by multiplying simpler distributions is also related~\cite{Hinton2002}.

Besides the network approaches discussed in the previous paragraph, the idea of partitioning the decision space and learning simpler decision functions in each partition has been explored. Mixture discriminant analysis treats each class as a mixture of Gaussians and learns discriminants between the Gaussians~\cite{Hastie1996}. Subclass discriminant analysis also relies on modeling classes as mixtures of Gaussians prior to learning discriminant~\cite{Zhu2006}. Local linear discriminant analysis clusters the data and learns a linear discriminant in each cluster~\cite{Kim2005}. In these approaches partitioning of the space is treated as a step independent from the supervised learning step. Wang and Saligrama, proposed a more recent approach that unifies space partitioning and supervised learning~\cite{Wang2012}. While this method is related in concept to our disjunctive learning, in Section~\ref{sec:resultsmulti} we show that LDNNs outperform space partitioning by a large margin. Dai {\em et al.} proposed an approach which places local classifiers close to the global decision boundary~\cite{Dai2006}. Toussaint and Vijayakumar propose a products-of-sigmoids model for discontinuously switching between local models~\cite{Toussaint2005}. Another approach greedily builds a piecewise linear classifier by adding classifiers in regions of error clusters~\cite{Dekel2012}. Local versions of SVMs have also been explored~\cite{Cheng2007,Hastie2001}. A specific type of local classification is based on the idea of pairwise coupling between positive and negative examples or clusters is conceptually close to the initialization we propose for our LDNN model. These methods typically employ a clustering algorithm, learning classifiers between pairs of positive and negative clusters found by clustering, finally followed by a combination scheme such as voting to integrate the pairwise classifiers into a single decision~\cite{Schulmeister1997,Hastie1998,Lu2004,Wu2004,Wu2007,Chen2008,Abbasi2012}. The modular network~\cite{Lu1999} discussed previously also falls into this category.

Despite the limitations mentioned earlier, artificial neural networks are the basis for highly successful Convolutional Neural Networks \cite{lecun3,lecun4} (ConvNet). ConvNets are special types of neural networks based on two properties: local connectivity and weight sharing. In general, a ConvNet consists of a few convolutional layers followed by one or more fully connected layers. The training process is done using error back-propagation all the way to the first layer. ConvNets have shown impressive results on many vision tasks including but not limited to classification, detection, localization and scene labeling \cite{lecun2, overfeat, mc, dropc, maxou}. They work best when large labeled data is available for training. For example, the state-of-the-art results for large 1000-category ’ImageNet’ \cite{imnet} dataset was significantly improved using ConvNets \cite{imagenet}. The main reason for this success is that ConvNets are strong feature learners for images. However, the classifier part in a ConvNet consists of one or a few layers of fully connected neural networks. These fully connected layers exhibit the limitations of MLPs including the herd-effect problem discussed earlier. There is a whole body of literature on the general task of classification in the past two decades that ConvNets simply cannot directly exploit in an unified structure because the classifier of choice is usually incompatible with learning by back-propagation and training the feature extractor and classifier independently does not lead to a structure with optimum performance. Usually, joint optimization of deep structures leads to a better solution or at least improves the result of layer-wise learning \cite{Hinton2006b}.

In this paper, we proposed a deep model which replaces fully connected layers with LDNN. We compared this model with state-of-the-art methods. These models are based on ConvNet. A notable and successful example is MCDNN proposed by Ciresan {\em et al.} \cite{mc}. In this model, they train multiple networks with slightly distorted and different inputs. The final classification is obtained by a voting scheme over probabilities of different networks.

The state-of-the-art results on many image classification datasets are achieved by DropConnect proposed by Wan {\em et al.} \cite{dropc}. The idea of DropConnect is inspired by Dropout \cite{HintonDropout}. They randomly drop the connections between the nodes instead of dropping output nodes of intermediate layers. We compared our model to ConvNets that use DropConnect in their fully connected layers. We show that our method is able to achieve competitive results with fewer epochs and smaller training parameters in most of the cases. Another recent example proposed by Goodfellow {\em et al.} is Maxout Networks \cite{maxou}. Instead of using an activation function over the output of a single node, they take the maximum output of a group of hidden nodes as the output. Here, the Max operator acts as an activation function. They also use a similar approach for convolutional layers. We provide comparisons with Maxout networks. But in general, they require significantly large networks because the output of a node or a convolutional map is determined by maximum of several input nodes or maps.


\section{Methods}
\label{sec:methods}
\subsection{Network Architecture}
\begin{figure}[h]
\begin{center}
\includegraphics[width=0.35\textwidth]{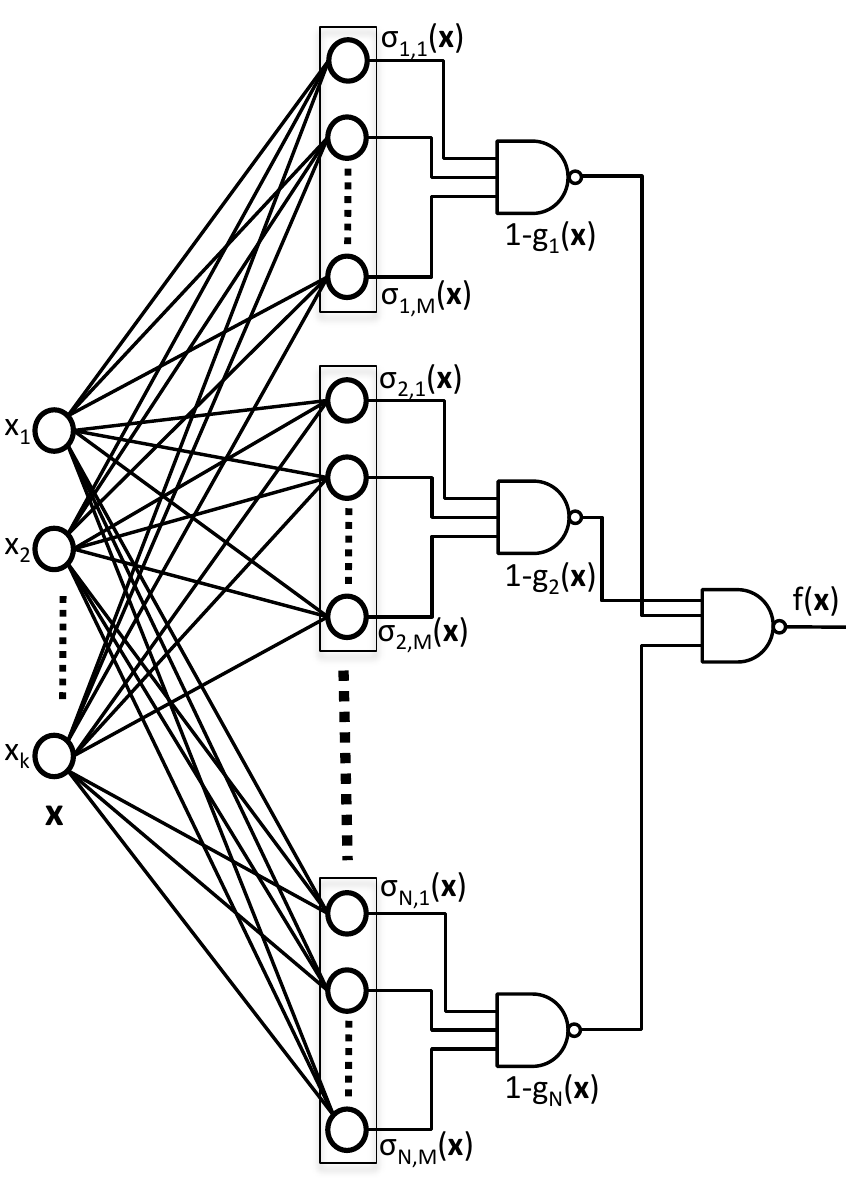}
\end{center}
\caption{\label{fig:arch}LDNN architecture. The first hidden layer is composed of $M\times N$ logistic sigmoid functions. The second hidden layer computes the logical negation of $N$ conjunctions using soft NAND gates. The output layer computes the disjunction. The soft NAND gates are implemented as continuous functions by subtracting the product of their inputs from 1.}
\end{figure}

Consider the binary classification problem $f:\mathbf{R}^n\rightarrow\mathbf{B}$ where $\mathbf{B}=\{0,1\}$.  Let $\Omega^+=\{\mathbf{x}\in \mathbf{R}^n:f(\mathbf{x})=1\}$. 
Lets approximate $\Omega^+$ as the union of $N$ convex polytopes $\tilde{\Omega}^+=\cup_{i=1}^{N} {\cal P}_i$ where the {\em i}'th polytope is the intersection ${\cal P}_i=\cap_{j=1}^{M_i} {\cal H}_{ij}$ of $M_i$ half-spaces ${\cal H}_{ij}=\{\mathbf{x}\in \mathbf{R}^n:h_{ij}(\mathbf{x})>0\}$. We can replace $M_i$ with $M=\max_i M_i$ without loss of generality. $H_{ij}$ is defined in terms of its indicator function 
\begin{equation}
h_{ij}(\mathbf{x})=\left\{ 
\begin{array}{lr}
1,& \sum_{k=1}^n w_{ijk}x_k+b_{ij}\geq 0\\
0,& {otherwise}
\end{array}
\right.,
\end{equation}
where $w_{ijk}$ and $b_{ij}$ are the weights and the bias term. Any Boolean function $b:\mathbf{B}^n\rightarrow\mathbf{B}$ can be written as a disjunction of conjunctions, also known as the disjunctive normal form~\cite{Hazewinkel2001}. Hence, we can construct the function
\begin{equation}
 \tilde{f}(\mathbf{x}) = \bigvee_{i=1}^N{ \underbrace{\left(\bigwedge_{j=1}^{M} h_{ij}(\mathbf{x})\right)}_{b_i(\mathbf{x})} }
\end{equation}
such that $\tilde{\Omega}^+=\{\mathbf{x}\in \mathbf{R}^n:\tilde{f}(\mathbf{x})=1\}$. Since $\tilde{\Omega}^+$ is an approximation to $\Omega^+$, it follows that $\tilde{f}$ is an approximation to $f$. Our next step is to provide a differentiable approximation to this disjunctive normal form. First, the conjunction of binary variables $ \bigwedge_{j=1}^{M} h_{ij}(\mathbf{x}) $ can be replaced by the product $\prod_{j=1}^M h_{ij}(\mathbf{x})$. Then, using De Morgan's laws~\cite{Hazewinkel2001} we can replace the disjunction of the binary variables  $\bigvee_{i=1}^N b_{i}(\mathbf{x}) $ with $\lnot\bigwedge_{i=1}^N\lnot{b_i(\mathbf{x})}$, which in turn can be replaced by the expression $1 - \prod_{i=1}^N (1- b_{i}(\mathbf{x}))$. Finally, we can approximate the perceptrons $h_{ij}(\mathbf{x})$ with the logistic sigmoid functions 
\begin{equation}
\sigma_{ij}(\mathbf{x}) = \frac{1}{1+e^{-\sum_{k=1}^n w_{ijk}x_k+b_{ij}}}.
\label{eq:sigmoid}
\end{equation}
This yields the differentiable approximation to $\tilde{f}$
\begin{equation}
\hat{f}(\mathbf{x}) = 1 - \prod _{i=1}^N 
(1-  
\underbrace{\prod_{j=1}^M \sigma_{ij}(\mathbf{x})
}_{g_i(\mathbf{x})}
),
\label{eq:ldnn}
\end{equation}
which can also be visualized as a network (Figure~\ref{fig:arch}). We refer to the proposed network architecture as LDNN. The only adaptive parameters of the LDNN are the weights and biases of the first layer of logistic sigmoid functions. The second layer consists of $N$ soft NAND gates which implement the logical negations of the conjunctions $g_i(\mathbf{x})$ using products. The output layer is a single soft NAND gate which implements the disjunction using De Morgan's law. We will refer to a LDNN classifier which has $N$ NAND gates in the second layer and $M$ discriminants per NAND gate as a $N\times M$ LDNN. Note that other variations of disjunctive normal networks can be constructed by using any classifier that is differentiable with respect to its parameters in place of the logistic sigmoid functions.

\subsection{Model Initialization}
\label{sec:init}
Consider a set of training examples $\Gamma=\{(\mathbf{x},y(\mathbf{x}))\}$ where $y(\mathbf{x})$ denotes the desired binary class corresponding to $\mathbf{x}$. Let $\Gamma^+$ and $\Gamma^-$ be the subsets of $\Gamma$ for which $y=1$ and $y=0$, respectively. The disjunctive normal form permits a very simple and intuitive initialization of the network weights. To initialize a $N\times M$ LDNN, we first partition $\Gamma^+$ and $\Gamma^-$ into $N$ and $M$ clusters, respectively. Let $\mathbf{v_{ij}} = \mathbf{c}^+_i - \mathbf{c}^-_j$ where $\mathbf{c}^+_i$ and $\mathbf{c}^-_j$ are the centroids of the i'th positive and j'th negative clusters, respectively.  We initialize the weight vectors as $\mathbf{w_{ij}} = \mathbf{v_{ij}}/|\mathbf{v_{ij}}|$. Finally, we initialize the bias terms $b_{ij}$ such that the logistic sigmoid functions $\sigma_{ij}(\mathbf{x})$ take the value $0.5$ at the midpoints of the lines connecting the positive and negative cluster centroids. In other words, let $b_{ij}=\langle\mathbf{w}_{ij},0.5(\mathbf{c}^+_i + \mathbf{c}^-_j)\rangle$ where $\langle\mathbf{a},\mathbf{b}\rangle$ denotes the inner product of the vectors $\mathbf{a}$ and $\mathbf{b}$.
This procedure initilizes $g_i(\mathbf{x})$, the {\em i}'th conjunction in the second hidden layer of the LDNN, to a convex polytope which aims to separate the training instances in the {\em i}'th cluster of $\Gamma^+$  from all training instances in $\Gamma^-$.  

\begin{figure}[t]
\begin{center}
\includegraphics[width=0.48\textwidth]{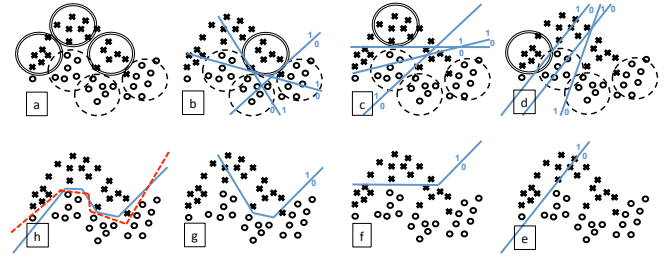}
\end{center}
\caption{\label{fig:init}A binary classification problem: (a) positive and negative training examples partitioned into three clusters each; linear discriminants from each negative cluster to (b) the first positive cluster, (c) the second positive cluster and (d) the third positive cluster; the conjunction of the discriminants for (g) the first positive cluster, (f) the second positive cluster and (e) the third positive cluster; (h) the disjunction of the conjunctions before (blue) and after (red) gradient descent. The 1/0 pair on the sides of the discriminants represent the direction of the discriminant.}
\end{figure}

We give an intuitive description of LDNN initialization in the context of the two moons dataset. An illustration of this dataset and three clusters for each of the two classes are shown in (Figure~\ref{fig:init}a). Initial discriminants for the positive clusters taken one at a time are shown in (Figure~\ref{fig:init}b-d). The conjunction of these discriminants form convex polytopes for the positive clusters (Figure~\ref{fig:init}e-g). The disjunction of these conjunctions before and after weight optimization (Section~\ref{sec:opt}) are illustrated in (Figure~\ref{fig:init}h). This initialization procedure is similar to the modular neural network proposed by Lu and Ito~{\em(12)} as well as to locally linear classification by pairwise coupling~{\em(20)} in general. Each module in Lu and Ito's modular network independently learns a linear classifier between a pair of positive and negative training data clusters. The key difference of our classifier from Lu and Ito's network, as well as from locally linear classification by pairwise coupling in general, is that we learn all the linear discriminants simultaneously by minimizing a single error function. When each module is trained independently, the success of the initial clustering can strongly influence the outcome. In Section~\ref{sec:results}, we show, using both real and artificial datasets, that this important disadvantage can create very significant differences in classification accuracy between modular networks and LDNNs.  

\subsection{Model Optimization}
\label{sec:opt}
The LDNN model can be trained by choosing the network weights and biases that minimize the quadratic error 
\begin{equation}
E({\cal W},\Gamma)=\sum_{(\mathbf{x},y)\in\Gamma} \left(y - f(\mathbf{x})\right)^2,
\label{eq:error}
\end{equation}
where $f$ is determined by the set of network weights and biases ${\cal W}$. Starting from an initialization as described in Section~\ref{sec:init}, we minimize (\ref{eq:error}) using gradient descent. To derive the update equations we need to find the partial derivatives of the error with respect to the network weights and biases. Using the fact that $\partial{\sigma_{ij}}/\partial{w_{pqk}}$ is non-zero only when $i=p$ and $j=q$, the derivatives of the error function with respect to the network weights is obtained using the chain rule 
\begin{align}
\label{eq:bw1}
\frac{\partial E}{\partial w_{ijk}} &= 
\frac{\partial E}{\partial f} 
\frac{\partial f}{\partial g_i} 
\frac{\partial g_i}{\partial \sigma_{ij}} 
\frac{\partial \sigma_{ij}}{\partial w_{ijk}} 
\nonumber \\
&=
-2(y- f(\mathbf{x}))
\left(\prod_{r\neq i}(1-g_r(\mathbf{x}))\right)\times
\nonumber \\
&~~~~~
\left(\prod_{l\neq j}\sigma_{il}(\mathbf{x})\right)
\left(\sigma_{ij}(\mathbf{x})(1-\sigma_{ij}(\mathbf{x}))x_k\right)
\nonumber \\
&=
2(f(\mathbf{x})-y)
\left(\prod_{r\neq i}(1-g_r(\mathbf{x}))\right)
g_i(\mathbf{x})
\left(1-\sigma_{ij}(\mathbf{x})\right)x_k
\end{align}
Similarly, we obtain the derivative of the error function with respect to the network biases as
\begin{equation}
\label{eq:bw2}
\frac{\partial E}{\partial b_{ij}} = 2(f(\mathbf{x})-y)
\left(\prod_{r\neq i}(1-g_r(\mathbf{x}))\right)
g_i(\mathbf{x})
\left(1-\sigma_{ij}(\mathbf{x})\right)
\end{equation}

We perform stochastic gradient descent after randomly permuting the order of the instances in $\Gamma$ and updating the model weights and biases according to 
$w_{ijk}^{new} = w_{ijk} - \alpha  \frac{\partial E}{\partial w_{ijk}}$, and $
b_{ij}^{new} = b_{ij} - \alpha  \frac{\partial E}{\partial b_{ij}}$,
respectively. The constant $\alpha$ is the step size. This constitutes one epoch of training. Multiple epochs are performed until convergence as determined using a separate validation set. Notice that it is possible to achieve $0$ training error for any finite training set $\Gamma$ by letting each positive training instance and each negative training instance represent a positive and negative cluster centroid, respectively. However, in practice, this is expected to lead to overfitting and poor generalization and typically a much smaller number of clusters than training instances is used. 

%
%

\subsection{Deep learning with LDNN}
\label{sec:conv}
\subsubsection{Convolutional feature learning}
We use $X*H$ for 2D convolution of $X$ and $H$, and $X\star H$ for 2D cross-correlation. We have the following equation for the forward pass of a convolutional layer:
\begin{equation}
X^{l}_{j}=\sigma(\underbrace{\sum_{i\in \mathbf{m}_j^l} X_i^{l-1}\star H_{ij}^{l} + b_j^l}_{S_j^l})
\label{eq:cnn1}
\end{equation}
In (\ref{eq:cnn1}), $l$ is the index of the convolutional layers and $i$ and $j$ are the indices of the layer maps. $X_i^{l-1}$ are the maps of the layer $l-1$ and $X_j^{l}$ are the maps of the layer $l$, $\mathbf{m}_j^l$ is the subset of maps in the layer $l-1$ that are connected to map $j$ of layer $l$ via filters $H_{ij}^l$. Finally, $\sigma$ is the activation function (e.g., logistic, ReLU). For the backward pass, we have the following sensitivity equations for the same convolutional layer:
\begin{align}
\frac{\partial E}{\partial S^{l}_{j}} &= \frac{\partial E}{\partial X^{l}_{j}}\circ \sigma^{\prime}(S^{l}_{j}),  \,\,\,\,
\frac{\partial E}{\partial H^{l}_{ij}} = X_{i}^{l-1} \star \frac{\partial E}{\partial S^{l}_{j}}, \nonumber \\ 
\frac{\partial E}{\partial b_j} &= \sum_{u,v} \frac{\partial E}{\partial S^{l}_{j}}, \,\,\,\,
\frac{\partial E}{\partial X^{l-1}_{j}} = \frac{\partial E}{\partial S^{l}_{j}} * H_{j}^{l}
\label{eq:cnn_upd}
\end{align}
$E$ is the error that we want to minimize. Here, the derivative of $E$ with respect to matrices $X$, $H$ and $S$ is a matrix consisting of derivatives of $E$ with respect to each element of that matrix. In the above equations $\circ$ is defined to be element-wise multiplication.

\subsubsection{proposed structure}
\begin{figure}[h]
\begin{center}
\includegraphics[width=0.49\textwidth]{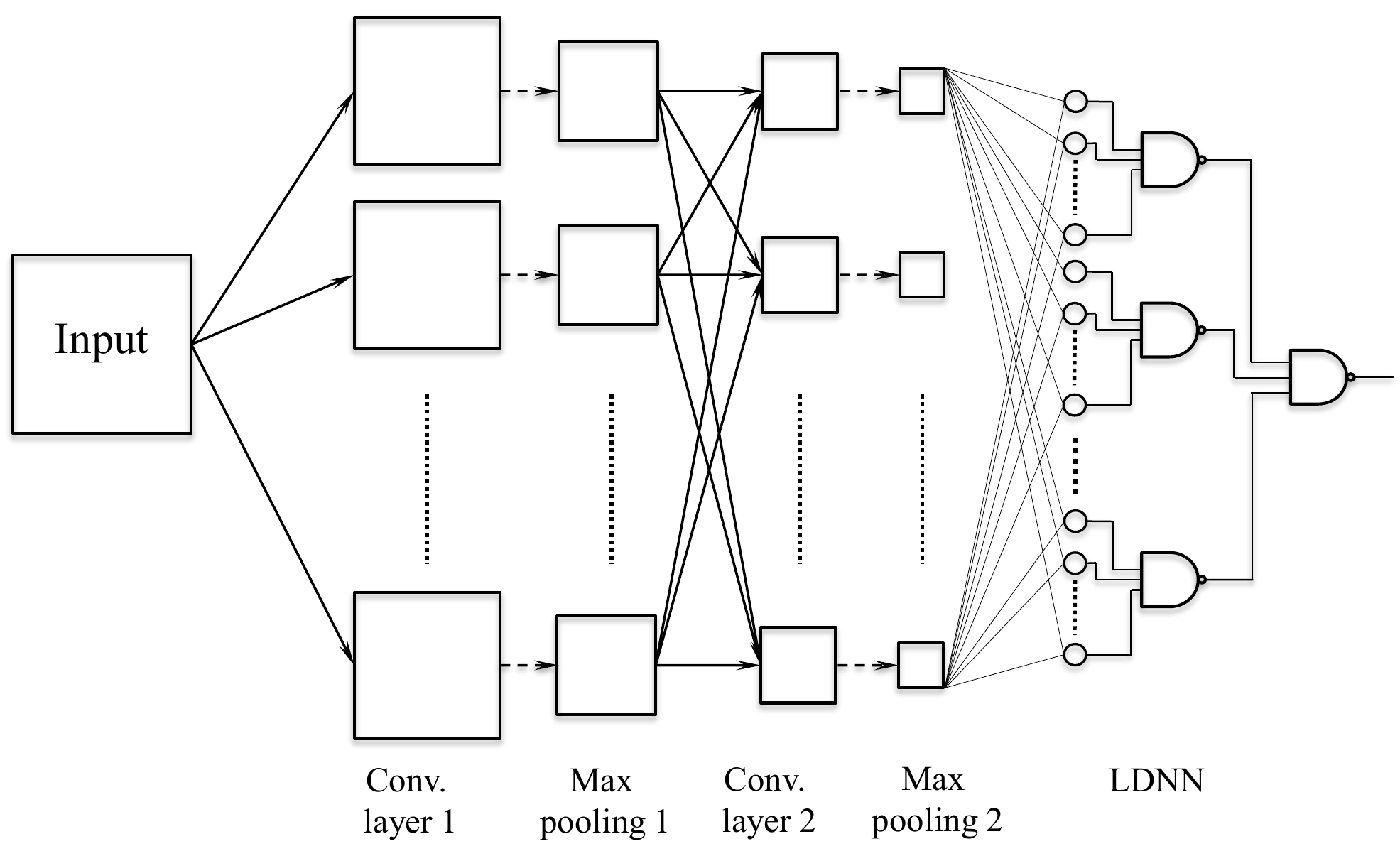}
\end{center}
\caption{\label{fig:struct}Proposed structure. Each solid arrow is a convolution of its input map with a 2D filter. Dashed arrows are Max pooling operators.}
\end{figure}

As mentioned earlier, LDNN consists of one layer of learnable weights and two layers of fixed soft gates. The error is back-propagated through soft gates to update the learnable weights. It is also possible to calculate the sensitivities for the input vector:
\begin{align}
\frac{\partial E}{\partial x_{k}} &= 
2(f(\mathbf{x})-y)
\sum_{i=1}^{N}\prod_{r\neq i}(1-g_r(\mathbf{x}))
g_i(\mathbf{x}) \times
\nonumber\\
&\sum_{j=1}^{M}\left(1-\sigma_{ij}(\mathbf{x})\right)w_{ijk},\,\,k=1,\hdots,n
\label{eq:LDNN_x1}
\end{align}
This allows us to train a convolutional feature extractor by back-propagating the error. In other words, we can seamlessly replace the fully-connected layers in ConvNet with LDNN. This combination is compatible because both ConvNet feature extractor and LDNN classifier are being trained using back-propagation via the chain rule. Our proposed structure is shown in Figure \ref{fig:struct}. 

For the case of multiclass classification, we use multiple LDNNs (one for every class). In this setup, the back-propagated errors for all the LDNNs will be summed together to form the sensitivities for convolutional layers. Assuming that $C$ is the number of classes:
\begin{align}
\frac{\partial E}{\partial x_{k}} &= 
\sum_{c=1}^{C} 2(f_c(\mathbf{x})-y_c)\sum_{i=1}^{N}\prod_{r\neq i}(1-g_{cr}(\mathbf{x}))
g_{ci}(\mathbf{x})
\nonumber \\
& \sum_{j=1}^{M}\left(1-\sigma_{cij}(\mathbf{x})\right)w_{cijk},\,\,k=1,\hdots,n
\label{eq:LDNN_x2}
\end{align}

$f_c(\mathbf{x})$ is output of the LDNN corresponding to class $c$. $y_c$ (for $c=1, \hdots, C$) is the label vector for datapoint $\mathbf{x}$. In the above equations, it is assumed that we are minimizing quadratic error. We can also minimize the cross-entropy loss function: $E({\cal W},\Gamma)=-\sum_{(\mathbf{x},y)\in\Gamma} y\log f(\mathbf{x}) + (1-y)\log(1-f(\mathbf{x}))$. It must be noted that using multiple LDNNs does not make the algorithm noticeably slower compared to a similarly configured ConvNet.

Assuming that we have $L$ convolutional layers, the forward pass for this structure starts from 
(\ref{eq:cnn1}). Beginning from input data, we use this equation recursively until we get the output maps of the last convolutional layer, which is $X_j^L$. The input for LDNN is formed by reshaping and concatenating the maps of the last convolutional layer into a vector: $\mathbf{x} = [\mathbf{x}_{1}^L, \mathbf{x}_{2}^L, \hdots, \mathbf{x}_{p(L)}^L]^T$. $p(L)$ is the number of maps in the last convolutional layer and $\mathbf{x}_{j}^{L}$ is the matrix $X_{j}^{L}$ reshaped into a vector. Then we can perform the forward pass of LDNN using (\ref{eq:ldnn}) and perform the backward pass using (\ref{eq:bw1}) and (\ref{eq:bw2}) to update LDNN weights and biases. Then we can use (\ref{eq:LDNN_x2}) to get sensitivities with respect to LDNN inputs: $[\partial E/\partial x_1, \partial E/\partial x_2, \hdots, \partial E\partial x_n]$. By reshaping this vector into 2D maps, we can get the sensitivities for maps in the last convolutional layer (i.e.,  $\partial E/\partial X^{L}_{j}$). Having these sensitivities, we can back-propagate the error to convolutional layers using (\ref{eq:cnn_upd}).

\section{Experiments on General Classification}
\label{sec:results}
\subsection{Artificial datasets}
\begin{figure}
\begin{center}
\begin{tabular}{cc}
\includegraphics[width=0.15\textwidth]{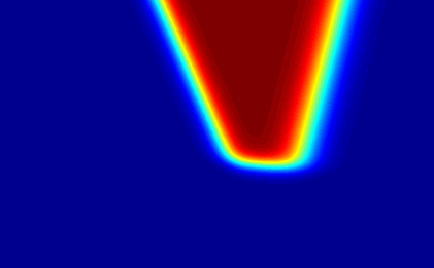}&
\includegraphics[width=0.15\textwidth]{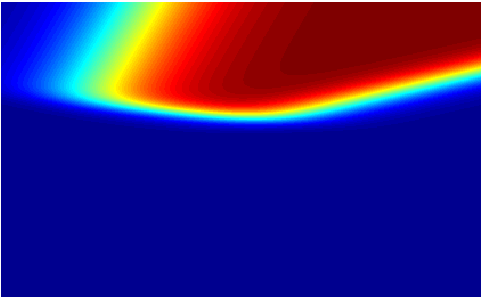}\\
(a) & (b)\\
\includegraphics[width=0.15\textwidth]{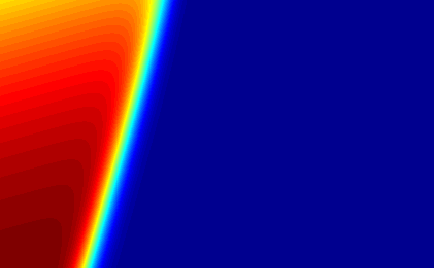}&
\includegraphics[width=0.15\textwidth]{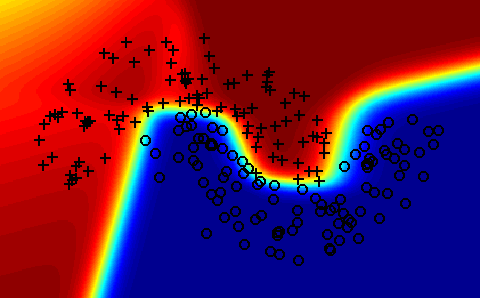}\\
(c) & (d)\\
\end{tabular}

\end{center}
\caption{\label{fig:moons}Two moons test set: (a)-(c) the 3 conjunctions in the second layer of the network evaluated individually, and (d) the output of the $3\times 3$ LDNN. +/o symbols denote the two classes.}
\end{figure}

We first experimented with the {\em two moons} artificial dataset to evaluate the LDNN algorithm with and without the proposed clustering initialization. We also compare the LDNN model with the modular neural networks(ModN))~\cite{Lu1999}. To construct the two moons dataset, we start by generating random radius and angle pairs $(r,\theta)$. For both moons, $r$ is an uniform random variable between $R-W/2$ and $R+W/2$ where $R$ and $W$ are parameters that determine the radius and the width of the moons, respectively. For the top moon, $\theta$ is an uniformly distributed random variable between $0$ and $\pi$. For the bottom moon, $\theta$ is an uniformly distributed random variable between $\pi$ and $2\pi$. The Cartesian coordinates for data points on the top and bottom moons are then generated as $(R\cos\theta,R\sin\theta)$ and $(R\cos\theta-W/2,R\sin\theta-\alpha)$, respectively. The parameter $\alpha$ determines the vertical separation between the two moons. We generated a training and a testing dataset by using the parameters $R=1$, $W=0.7$, $\alpha=-0.7$ which generates slightly overlapping classes. Both datasets contained 1000 instances on the top moon and 1000 instances on the bottom moon. Then, for each $n\in[1,7]$, we trained 50 $n\times n$ LDNNs starting from random parameter initializations, 50 $n\times n$ LDNNs initialized from k-means clustering with $n$ clusters per moon and 50 $n\times n$ ModNs initialized from k-means clustering with $n$ clusters per moon. For ModNs, the $n^2$ linear discriminants are trained independently using data from the $n^2$ pairs of positive (top moon) and negative (bottom moon) clusters and then combined using min/ma functions. We used stochastic gradient descent with a step size of $0.3$, a momentum term weight of $0.1$ and $500$ epochs for training all models. Testing accuracies were computed over the second dataset which was not used in training. Table~\ref{table:two-moons} shows the mean, minimum and maximum testing error over the 50 trials for each of the models. We observe that training the LDNN model starting from a random initialization is successful in general; however, the range of testing error rates varies by a larger amount compared to when a cluster initialization is used resulting in a slightly worse mean testing error. We also note that the LDNN model performs better both on average and when comparing the maximum error rates over the 50 trials than the ModN model. Figure~\ref{fig:moons} illustrates the output of the LDNN model for $n=3$, which appears to be an appropriate choice based on Table~\ref{table:two-moons}. The outputs of the $3$ conjunctions are also shown separately to give further intuition into the behavior of the LDNN model. Notice the similarity to Figures~\ref{fig:init}(e-h)).

 \begin{table}
 \begin{center}
 \begin{tabular}{|l|c|c|c|c|c|c|}
 \hline
 n &  \multicolumn{2}{|c|}{LDNN random init} &  \multicolumn{2}{|c|}{LDNN cluster init} &  \multicolumn{2}{|c|}{ModN cluster init} \\   
 & Av. &  Range & Av. & Range & Av. & Range \\
 \hline
1 & 15.6 & [15.2, 18.6] & 15.6 & [15.2, 20.2] & 15.5 & [15.2, 16.3]\\
\hline
2 & 6.6 & [3.0, 15.8] & 3.3 & [2.9, 3.7] & 4.2 & [3.6, 5.4]\\
\hline
3 & 4.1 & [1.1, 15.6] & 2.3 & [1.2, 3.5] & 2.7 & [1.2, 4.8]\\
\hline
4 & 3.6 & [1.2, 15.6] & 2.2 & [1.3, 3.5] & 3.0 & [1.8, 5.2]\\
\hline
5 & 3.4 & [1.2, 15.4] & 2.2 & [1.2, 4.2] & 2.8 & [1.4, 5.7]\\
\hline
 \end{tabular}
 \end{center}
 \caption{\label{table:two-moons} Average, min. and max. testing error percentages over 50 repetitions for LDNN initialized with random parameters, initialized with clustering and ModN~\cite{Lu1999} initialized with clustering for different model sizes.}
 \end{table}

\begin{figure}[htb]
\begin{center}
\begin{tabular}{ccc}
\includegraphics[width=0.14\textwidth]{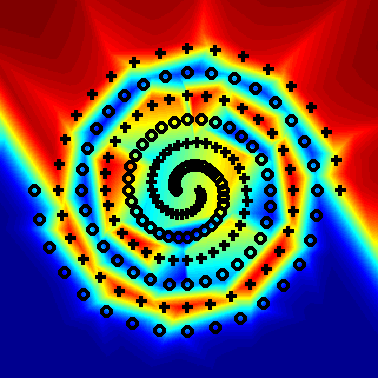}&
\includegraphics[width=0.14\textwidth]{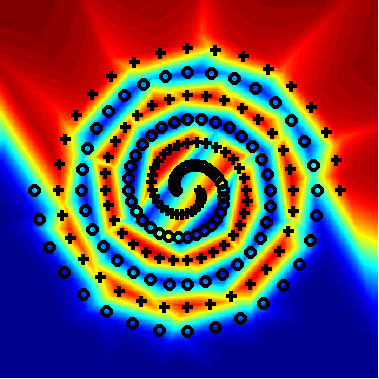}&
\includegraphics[width=0.14\textwidth]{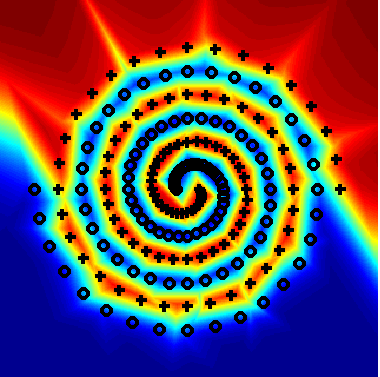}\\
\small{Training er. = 23\%} & {\small Training er. = 6\%} & \small{Training er. = 0\%}\\
\includegraphics[width=0.14\textwidth]{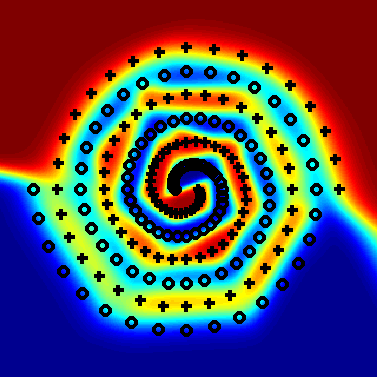} &
\includegraphics[width=0.14\textwidth]{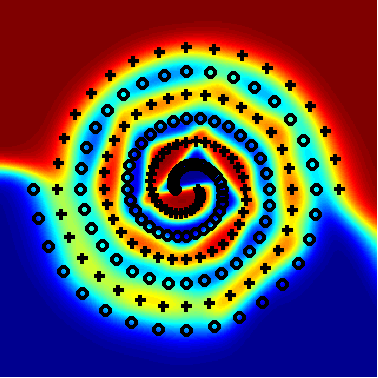} &
\includegraphics[width=0.14\textwidth]{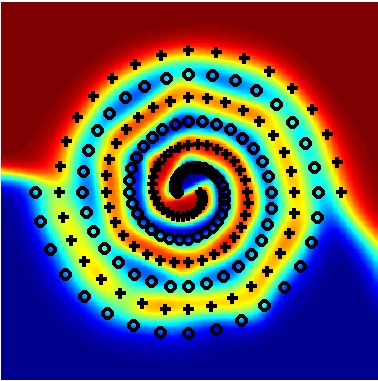}\\
\small{Training er. = 0\%} & {\small Training er. = 0\%} & \small{Training er. = 0\%}\\
{\bf 18 clusters/class} & {\bf 21 clusters/class} & {\bf 27 clusters/class}
\end{tabular}
\end{center}
\caption{\label{fig:spirals}Two spirals dataset: ModN (top) and LDNN (bottom).}
\end{figure}

The {\em two-spirals} dataset is an extremely difficult dataset for the MLP architecture trained with the backpropagation algorithm~\cite{Fahlman90thecascade-correlation}. The original dataset consists of 194 $(x,y)$ pairs arranged in two interlocking spirals that orbit the origin three times. The classification task is to determine which spiral any given $(x,y)$ point belongs to. We used the farthest distance clustering algorithm~\cite{DudaHart} for initialization of both models. The k-means clustering algorithm places most centroids near the origin where the data points are denser and fewer centroids on the spiral arms further from the origin where the data is sparser. On the other hand, the farthest distance clustering algorithm provides more uniformly distributed centroids which leads to better classification results with fewer clusters. We performed clustering with maximum distance thresholds $2.2$, $2.0$ and $1.5$ resulting in $18$, $21$ and $27$ clusters per class, respectively. For each of these, we trained a LDNN and a ModN. Note that the number of parameters in both models is the same for the same number of clusters. We used stochastic gradient descent with a step size of $0.3$, a momentum term weight of $0.1$ and $2,000$ epochs for training all models. LDNN achieved $0$ percent training error in each of these cases while the ModN's training error was $0.232$, $0.062$ and $0$ percent, respectively. These results suggest that the unified learning framework of LDNN is able to capture the spiral dataset with many fewer parameters than independent, pairwise learning of discriminants as in~\cite{Lu1999}. Furthermore, it can be seen from Figure~\ref{fig:spirals} that LDNN creates a much smoother approximation to the spirals than pairwise learning. Finally, we note that LDNN initialized randomly was not able to find a satisfactory local minimum of the error function via gradient descent. This is similar to the failure of the standard MLP architecture for this dataset. This observation underlines the importance of the existence of an intuitive initialization for the LDNN architecture. 

\subsection{Two-class problems}
\label{sec:binary}
We experimented with 10 different binary classification datasets from the {\em UCI Machine Learning Repository}~\cite{uci} and the {\em LIBSVM Tools} webpage~\cite{libsvm}. For each dataset, we trained the LDNN, ModN, MLP, SVM and RF classifiers. 
\subsubsection{Dataset normalization, training/testing set split}
\label{sec:normalize}
Datasets were normalized as follows: For LDNN, ModN and MLP training, we applied a whitening transform~\cite{DudaHart} to datasets with a large number of training instances (Forest cover type and Webspam) since the covariance matrix could be estimated reliably. All other datasets were normalized by centering each dimension of the feature vector at the origin by subtracting its mean and then scaling by dividing it with its standard deviation. For SVM training, each dimension of the feature vector was linearly scaled to the range $[0,1]$. For RF training, no normalization is necessary. 

The IJCNN and COD RNA  binary datasets had previously determined training and testing sets. For the rest of the datasets, we randomly picked 2/3's of the instances for training and the rest for testing. For LDNN, MLP, MLP-m and Mod-N experiments, the training set was further randomly split into a training (\%90) and cross-validation (\%10) set for determining the number of epochs to use in training. 

\subsubsection{Model and classifier training parameter selection}
\label{sec:modelchoose}
For LDNN classifiers we need to choose the number of NAND gates (N) and the number of discriminants per group (M). These parameters translate into the number of positive and negative clusters, respectively in the initialization. Various combinations, up to 40 clusters per class, were tried to find the selection that gives the best testing accuracy. For any given number of clusters, the k-means algorithm was repeated 50 times and the clustering result with the lowest sum of square distances to nearest cluster centroid was selected to initialize the LDNN weights. We also fine tuned the step size for gradient descent. The number of epochs for training was selected using the cross-validation set except for the IJCNN dataset. For the IJCNN dataset cross-validation set was also used in training as in~\cite{Chang2001} and the number of epochs was fixed at 20.

For MLP training, there are two main parameters. The first one is the number of hidden nodes which was varied from $2$ to $40$ to find the best test set accuracy. This was followed by fine tuning the step size for backpropagation. The number of epochs was chosen using the cross-validation set.  We also trained a second MLP classifier (MLP-m) for which the number of hidden nodes was chosen as $N\times M$ to match the total number of logistic sigmoid functions in the LDNN classifier. This was done to compare LDNN to a MLP with approximately the same degrees of freedom. It was not feasible to train the MLP-m classifier for 4 of the datasets due to extremely long training times. Similarly, a modular network, which we refer to as Mod-N, with the same number of conjunctions and disjunctions as the LDNN classifier was trained to control for the degrees of freedom.

There are three main parameters involved in RF training. The first one is the number of trees. We choose a sufficiently large number of trees to ensure that the out of bag error rate converges. The second parameter is the number of features that will be considered in every node of the tree. We tried a range of numbers around the square root of the number of features~\cite{Breiman2001}. The last parameter is the fraction of total samples that will be used in the construction of each tree. We tried $2/3$, $1/2$, $1/3$, $1/4$ and $1/5$ as possible values for this parameter.

For SVM training, a RBF kernel was used for all datasets except for the MNIST dataset for which a 9th degree polynomial kernel was used~\cite{Hinton2006b}. For all datasets except MNIST, we used the grid search tool provided by the Libsvm guide~\cite{libsvm} to set the parameters of the RBF kernel.

The training and model parameters selected for all models are listed in Table~\ref{table:binaryresults}. 

\begin{table*}[!b]
\begin{center}
\begin{tabular}{|l||c|c|c|c|c|l|}
\hline
 Dataset, source and properties & Classifier & Av. train err & Av. test err & Test err range & Time & Model Parameters\\
\hline
{\em Adult} & LDNN & 15.13 & 15.25 & [15.14, 15.41] & 3.1 &  $7\times 4$, $\epsilon=0.007$ \\
UCI & MLP & 15.24 & 15.75 & [15.27, 15.95] & 90.6 &  $h=10$, $\epsilon=0.0005$ \\ 
Train: 7,508+ / 22,654-& RF & 7.28 & {\bf 14.14} & [13.97, 14.30] & 9.3 &  $t=300$, $f=3$, $s=2/3$\\ 
 Test: 3,700+ / 11,306- & SVM & 15.41 & 15.57 &---& 162.7 &  $C=32768$, $\gamma=0.007812$\\
Dim = 14 & Mod-N & 17.38 & 17.39 & [16.32, 20.51] & 0.9 &  $7\times 4$, $\epsilon=0.007$\\ 
& MLP-m & 14.78 & 15.81 & [15.27, 16.32] & 175.3 &  $h=28$, $\epsilon=0.0005$ \\
\hline
{\em Wisconsin breast cancer} & LDNN & 1.95 & {\bf 0.80} & [0.52, 1.58] & $<$0.1&  $2\times 1$, $\epsilon=0.05$ \\
UCI & MLP & 1.79 & 2.28 & [0.52, 3.70] & 0.9 &  $h=4$, $\epsilon=0.001$ \\
Train: 142+ / 238-& RF & 0.32 & 1.79 & [1.58, 2.11] & $<$0.1&  $t=300$, $f=10$, $s=2/3$ \\
Test: 70+ / 119-& SVM & 2.63 & 1.59 &--- & $<$0.1&  $C=2048$, $\gamma=0.000488$\\
Dim = 30 &Mod-N & 15.58 & 14.58& [7.93, 24.33]& $<$0.1&  $2\times 1$, $\epsilon=0.05$  \\
&MLP-m & 2.22 & 2.29 & [1.05, 4.23] & 1.1 &  $h=2$, $\epsilon=0.0005$ \\
\hline
{\em PIMA Indians diabetes} & LDNN & 20.94 & {\bf 17.92} & [17.25, 19.60] & 0.2 &  $6\times 10$, $\epsilon=0.02$\\
UCI & MLP & 22.19 &  22.11 & [19.60, 24.31] & 0.4&  $h=6$, $\epsilon=0.001$\\
Train: 179+ / 334-& RF & 13.20 & 20.81 & [20.39, 21.56] & $<$0.1&  $t=150$, $f=2$, $s=1/5$ \\ 
Test: 89+ / 166-  & SVM & 21.83 & 21.57 &--- & $<$0.1&  $C=32$, $\gamma=0.125$ \\
Dim = 8 & Mod-N & 20.11 &24.29 &[19.60, 27.05]& $<$0.1&  $6\times 10$, $\epsilon=0.02$ \\ 
& MLP-m & 18.79 & 24.98 & [18.43, 31.76]& 1.7 &  $h=60$, $\epsilon=0.0005$ \\
\hline
{\em Australian credit approval} & LDNN & 10.04 & {\bf 12.93} & [12.22, 13.53] & $<$0.1&  $5\times 4$, $\epsilon=0.02$\\
UCI & MLP & 9.77 & 15.65 & [13.10, 18.34] & 0.5&  $h=8$, $\epsilon=0.001$\\
Train: 205+ / 256-&RF & 10.95 & 12.95 & [12.22, 13.10] & $<$0.1&  $t=150$, $f=1$, $s=1/5$\\
Test: 1012+/127- & SVM & 13.02 & 16.59 &--- & $<$0.1&  $C=0.03125$, $\gamma=0.5$\\
Dim = 14& Mod-N& 10.43 & 14.62 & [12.22, 17.46]& 1.0 &  $5\times 4$, $\epsilon=0.02$ \\
&MLP-m & 8.29& 16.74 & [13.53, 20.08]& $<$0.1 &  $h=20$, $\epsilon=0.001$ \\
\hline
{\em Ionosphere} & LDNN & 1.28 & {\bf 3.40} & [2.56, 4.27]  & 0.2 &  $1\times 36$, $\epsilon=0.05$\\
UCI & MLP & 4.80 & 12.10 & [6.83, 20.51] & 0.4 &  $h=6$, $\epsilon=0.005$\\
Train: 150+ / 84-& RF & 5.42 & 5.38 & [5.12, 5.98] & $<$0.1 &  $t=200$, $f=5$, $s=1/5$\\ 
Test: 75+ / 42-& SVM & 0.85 & 4.27 &--- & $<$0.1 &  $C=2$, $\gamma=2$\\
Dim = 33& Mod-N & 1.74 & 5.98 &[4.27, 8.54]& $<$0.1 &  $1\times 36$, $\epsilon=0.05$ \\ 
 & MLP-m & 3.05 & 9.47 & [5.98, 15.38]& 2.7 &  $h=36$, $\epsilon=0.005$ \\
\hline
{\em German credit} & LDNN & 17.54 & {\bf 22.58} & [21.02, 23,42] & 0.2 &  $6\times 1$, $\epsilon=0.05$\\
UCI & MLP & 19.85 & 26.96 & [22.52, 30.93] & 0.8&  $h=6$, $\epsilon=0.001$\\
Train: 200+ / 467-& RF & 1.07 & 24.28 & [23.42, 24.92] & 0.2&  $t=250$, $f=4$, $s=2/3$\\
Test: 100+ / 233-& SVM & 11.09 & 25.83 &--- & $<$0.1 &  $C=8$, $\gamma=0.125$\\
Dim = 24& Mod-N & 29.98 & 30.03 & [30.03, 30.03]& $<$0.1 &  $6\times 1$, $\epsilon=0.05$ \\
 & MLP-m & 19.85 & 26.92 & [22.52, 30.93]& 0.8 &  $h=6$, $\epsilon=0.001$ \\
 \hline
{\em Forest cover type} & LDNN & 8.22 & 8.87 & [8.09, 9.96] & 2702 &  $20\times 10$, $\epsilon=0.1$ \\
UCI & MLP & 11.76 & 12.22 & [11.45, 13.31] & 7499&  $h=40$, $\epsilon=0.001$ \\ 
Train: 188,868+ / 198,474-& RF & 0.29 & {\bf 3.90} & [3.84, 3.94] & 571.1&  $t=150$, $f=15$, $s=2/3$ \\ 
Test: 94,443+ / 99,237- & SVM & 6.03 & 6.91 &--- & 13043 &  $C=32$, $\gamma=8$ \\
Dim = 54& Mod-N & 25.52 & 25.68  & [24.40, 26.77] & 55.1 &  $20\times 10$, $\epsilon=0.1$\\
\hline
{\em IJCNN challenge} & LDNN & 0.87 & {\bf 1.28} & [1.02, 1.58] & 8.2&  $10\times 8$, $\epsilon=0.25$\\
Libsvm & MLP & 1.19 & 2.34 & [1.77, 3.08] & 294.8 &  $h=20$, $\epsilon=0.001$\\
 Train: 3,415+ 31,585-& RF & 0.08 & 2.00 & [1.91, 2.09] & 18.7&  $t=250$, $f=3$, $s=2/3$\\
Test: 8,712+ / 82,889- & SVM & 0.30 & 1.41 &--- & 38.4 &  $C=32$, $\gamma=8$\\
Dim = 22& Mod-N & 4.68 & 5.01  & [4.13, 7.95]& 2.7 &  $10\times 8$, $\epsilon=0.25$\\
\hline
{\em COD-RNA} & LDNN & 3.59 & {\bf 3.36} & [3.30, 3.46] & 80.9&  $8\times 8$, $\epsilon=0.05$  \\
Libsvm & MLP & 4.30 & 3.68 & [3.43, 4.02] &  168.3 &  $h=15$, $\epsilon=0.001$ \\
Train: 19,845+ / 39,690-& RF & 0.34 & 3.37 & [3.34, 3.39] & 8.3&  $t=200$, $f=3$, $s=2/3$ \\
 Test: 90,539+ / 181,07-& SVM & 2.86 & 3.67 &--- & 157.6&  $C=512$, $\gamma=8$ \\ 
Dim = 8 & Mod-N & 5.29 & 4.15 & [2.82, 4.72]& 2.0 &  $8\times 8$, $\epsilon=0.05$\\
\hline
{\em Webspam} & LDNN & 0.51 & 1.21 & [1.12, 1.27] & 401.8 &  $15\times 15$, $\epsilon=0.1$\\
Libsvm & MLP & 1.42 & 2.44 & [2.28, 2.60] & 6350&  $h=20$, $\epsilon=0.005$ \\
Train: 141,460+ / 91,874-& RF & 0.02 & 1.17 & [1.13, 1.19] & 428.0&  $t=100$, $f=11$, $s=2/3$\\
Test:  70,729+ / 45,937- & SVM & 0.30 & {\bf 0.78} &--- & 5345&  $C=8$, $\gamma=8$\\
Dim = 138 & Mod-N & 4.52 & 4.57 & [3.89, 5.17]& 67.7 &  $15\times 15$, $\epsilon=0.1$ \\
\hline
\end{tabular}
\end{center}
\caption{\label{table:binaryresults}
{\bf Column 1}: Binary classification datasets, their source, number of positive/negative training/testing examples and data dimensionality. 
{\bf Column 2}: Classifier type. 
{\bf Column 3-6}: Average training, average testing, [min,max] testing error (\%)  and computation time (seconds). Computation times less than $0.1$ seconds are not reported. Best average testing errors are shown in bold.
{\bf Column 7}: Model and classifier training parameters used. LDNN, Mod-N: $N\times M$ model and $\epsilon$ step size. MLP and MLP-m: $h$ number of hidden nodes and $\epsilon$ step size. RF: $t$ number of trees, $f$ number of features considered per node and $s$ training instance sampling rate for each tree. SVM: $C$ penalty factor, $\gamma$: RBF kernel width.}
\end{table*}

\subsubsection{Results}
All of the classifiers we consider, with the exception of SVM, are stochastic. Therefore, each experiment with the exception of SVM was repeated $50$ times to obtain mean, minimum and maximum testing errors which are reported in Table~\ref{table:binaryresults} for all classifiers.  The LDNN classifier outperformed MLPs for all datasets. Furthermore, LDNNs also outperform MLP-m in all datasets for which the MLP-m classifier was trained. In 8 out of 10 datasets, the mean LDNN error was smaller than the minimum MLP error, and, in 5 out of 6 datasets, the mean LDNN error was smaller than the minimum MLP-m error. All algorithms were run on an {\em Intel i7-3770} 3.4 Ghz CPU. In all datasets LDNNs were significantly faster to train than the MLPs. These results signify that the LDNN network architecture and training offers a faster to train and more accurate alternative to MLPs and backpropagation. The LDNN classifier also outperformed the Mod-N classifier in all datasets including several datasets such as {\em Forest cover type} and {\em Wisconsin breast cancer} where the accuracy difference was very large. This emphasizes the importance of training the entire network in an unified manner. Considering all of the classifiers tested, LDNNs had the lowest testing error in 7 out of 10 datasets. LDNNs outperformed SVMs in 8 out of 10 cases and RFs in 7 out of 10 cases. In 5 out of 10 cases the mean LDNN error was lower than the minimum RF error. The RF mean error was lower than the LDNN minimum error in only 2 out of 10 cases. Finally, LDNNs never severely over fit the data, whereas RFs  has significant accuracy differences between training and testing sets for several datasets including {\em Adult}, {\em PIMA Indian diabetes}, {\em German credit} and {\em Forest cover type}. 

\subsection{Multi-class problems}
\label{sec:resultsmulti}
\begin{table*}[!t]
\begin{center}
\begin{tabular}{|l||c|c|c|c|l|}
\hline
Dataset, source and properties & Classifier & Av. train err & Av. test err & Test err range & Model Parameters\\
\hline
{\em Isolet} & LDNN & 0.25 & 4.17 & [3.65, 4.49] & $4\times 4$, $\epsilon=0.01$ \\
Train: 6,238 / Test: 1,559 & RF & 0 & 5.61 & [5.25, 5.90] & $t=200$, $f=30$, $s=2/3$ \\
 Classes=26, Dim=617& SVM & 0 & {\bf 3.21} & ---  & $C=8$, $\gamma=0.03125$ \\
 & SP-LDA & ---& 5.58 & --- & Results taken from~\cite{Wang2012} \\
\hline 
{\em Landsat} & LDNN & 2.66 & {\bf 7.98} & [7.65, 8.25] &  $9\times 9$, $\epsilon=0.1$ \\
Train: 4,435 / Test: 2,000 & RF & 0.22 & 9.15 & [8.65, 9.55] &  $t=200$, $f=6$, $s=2/3$ \\
Classes=6, Dim=36 & SVM & 1.98 & 8.15 & --- & $C=2$, $\gamma=8$\\
 & SP-LDA & --- & 13.95 & --- & Results taken from~\cite{Wang2012} \\
\hline
{\em Letter} & LDNN & 0.20 & {\bf 2.32} & [2.12, 2.72] &  $20\times 20$, $\epsilon=0.4$ \\
Train: 16,000 / Test: 4,000 & RF & 0 & 3.89 & [3.65, 4.02] &  $t=500$, $f=3$, $s=2/3$ \\
Classes=26, Dim=16  & SVM & 0.08 & 2.35 & --- & $C=8$, $\gamma=8$ \\
  & SP-LR & --- & 13.08 & --- & Results taken from~\cite{Wang2012} \\
\hline
{\em Optdigit} & LDNN & 0.01 & 2.29 & [2.00, 2.67] &  $5\times 5$, $\epsilon=0.1$ \\
Train: 3,823 / Test: 1,797 & RF & 0 & 2.89 & [2.50, 3.11] &  $t=200$, $f=7$, $s=2/3$ \\
Classes=10, Dim=62 & SVM & 0.03 & {\bf 1.56} &--- &  $C=8$, $\gamma=0.125$\\
 & SP-P & ---& 4.23 & ---& Results taken from~\cite{Wang2012}  \\
\hline
{\em Pendigit} & LDNN & 0.34 & {\bf 1.80} & [1.68, 1.94] &  $8\times 8$, $\epsilon=0.005$ \\ 
Train: 7,494 / Test: 3,498 & RF & 0.01 & 3.64 & [3.40, 3.83] &  $t=250$, $f=4$, $s=2/3$\\
 Classes=10, Dim=16& SVM & 0.03 & 1.86 & --- &  $C=8$, $\gamma=2$ \\
 & SP-P & --- & 4.32 &--- & Results taken from~\cite{Wang2012} \\
\hline
{\em MNIST} & LDNN & 0.03 & {\bf 1.23} & [1.23, 1.23] &  $30\times 30$, $\epsilon=0.45$\\
Train: 60,000 / Test: 10,000 & RF & 0 & 3.00 & [2.88, 3.14]  &  $t=500$, $f=26$, $s=2/3$\\
 Classes=10, Dim=717& SVM & --- & 1.40 & --- &  Results taken from~\cite{Hinton2006b} \\
\hline
\end{tabular}
\end{center}
\caption{\label{table:multiresults}
{\bf Column 1}: Multi-class datasets, their source, number of training/testing examples and data dimensionality. 
{\bf Column 2}: Classifier type. 
{\bf Column 3-5}: Average training, average testing, and [min,max] testing error (\%). Best average testing errors are shown in bold.
{\bf Column 6}: Model and classifier training parameters used. LDNN: $N\times M$ model and $\epsilon$ step size. RF: $t$ number of trees, $f$ number of features considered per node and $s$ training instance sampling rate for each tree. SVM: $C$ penalty factor, $\gamma$: RBF kernel width. The space partitioning (SP) results are from~\cite{Wang2012}.}

\end{table*}

We also experimented with  6 multi-class datasets from the {\em UCI Machine Learning Repository}~\cite{uci}. Each dataset was first normalized in the same way as described in Section~\ref{sec:normalize}. For each dataset we trained the LDNN, RF and SVM classifiers with the exception of the MNIST dataset for which the SVM results are reported from~\cite{Hinton2006b}. In that paper, a SVM is trained on a feature set generated by a deep belief network. The model and classifier training parameters were chosen as described in Section~\ref{sec:modelchoose} and are reported in Table~\ref{table:multiresults}. 
LDNN and RF experiments were repeated $20$ times to obtain mean, minimum and maximum testing errors which are reported in Table~\ref{table:multiresults}.
The LDNN classifier is also related to the idea of space partitioning~\cite{Wang2012} which combines partitioning of the space and learning a local classifier for each partition into a global objective function for supervised learning. All space partitioning classifier results are reported from~\cite{Wang2012}. LDNNs had the best accuracy in 4 out of 6 datasets. Note that the minimum and maximum testing errors for LDNNs were equal for MNIST.

\section{Experiments on Image Classification}
In this section, we evaluate the performance of our proposed deep structure through different experiments. We incorporated our LDNN into a GPU implementation of ConvNet that is publicly available at \cite{cudac}. Most of the experiments described here are done using this GPU implementation. In all the experiments, we found good parameters using cross-validation on a small portion of training data and repeat the training on all training data. For all the datasets, we repeated the experiments a number of times and report two  numbers here. One is the average over error rates of different experiments and the other is the voting result of different experiments. For obtaining the voting result, we average the predicted probabilities of different experiments before calculating the error rate. Basically the only difference between multiple runs of the experiments for every dataset is the random initialization of weights.
\subsection{MNIST}
MNIST \cite{lecun4} is probably the most popular dataset in the area of digit classification. It contains 60000 training and 10000 test samples of
size 28$\times$28 pixels. We conducted two sets of experiments on this dataset: the first on unmodified MNIST data and the second on the MNIST 
with the augmented training set. 
\begin{figure}[h]
\begin{center}
\includegraphics[width=0.44\textwidth]{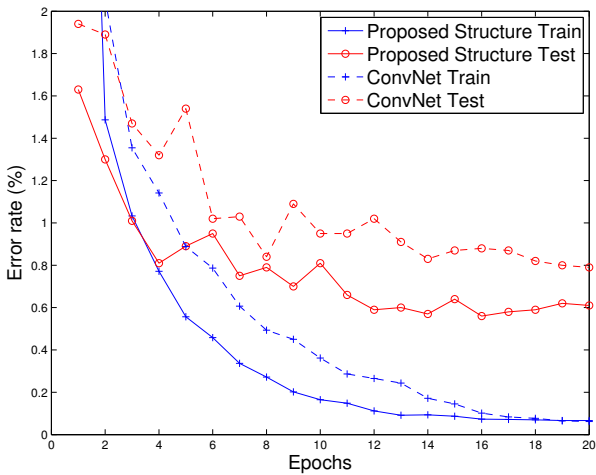}
\end{center}
\caption{\label{fig:converg}Convergence of our method compared to conventional ConvNet on train and test data of MNIST.}
\end{figure}
For the first task, we used 2 convolutional layers as feature detectors. The first layer uses 7$\times$7 filters and produces 20 maps. The second layer also uses 7$\times$7 
filters but produces 15 maps. Ten LDNNs perform the classification part (one per class). Every LDNN consists of 5 groups and 5 discriminants per group. No preprocessing was performed on this dataset. We trained 4 networks with this setting and used voting for the final classification. We did not use annealing, momentum or any type of regularization 
for this case. However, the convergence is very fast and on average it requires less than 20 epochs, significantly faster than any other 
neural network-based method, which require hundreds of epochs to train \cite{dropc,stoch}. Furthermore, the number of trainable parameters in our case is less than in most 
other ConvNet-based methods.
Using this setting, we obtained 0.38\% test error rate on the unmodified MNIST dataset. Table \ref{tab1} compares this error rate to other algorithms. The solutions based on maxout Networks \cite{maxou} 
(0.45\%) and stochastic pooling \cite{stoch} (0.47\%) also use convolutional feature learning but require more trainable parameters or training epochs. For example, stochastic pooling uses 3 convolutional layers with 64 maps in each layer and 280 training epochs. Results of Maxout Networks are obtained by using 3 convolutional layers and 96 maps in each of them. In a matched experiment, we compared the convergence properties of our method to  conventional ConvNet. The number of nodes in fully-connected layer of ConvNet is set to be the same as the number of linear discriminants in LDNN. For ConvNet, another layer of weights was added to connect the fully-connected layer to ten outputs. Everything else is exactly the same for both structures. The convergence plot is shown in Figure \ref{fig:converg}. We can see that our structure performs better in terms of accuracy and convergence. It must be noted that a conventional ConvNet is significantly faster to converge compared to structures with Dropout or DropConnect.
\begin{table}[h]  
  \begin{center}
  \begin{tabular}{ | c | c | }
  \hline
                 & Error rates: \\
    Method         & Voting (\%) ( Average (\%) $\pm$ std. dev)\\ \hline
    Proposed structure   & 0.38 (0.60 $\pm$ 0.027) \\
    DropConnect \cite{dropc}    & 0.57 (0.63 $\pm$ 0.035) \\
    Dropout \cite{dropc} & 0.52 (0.59 $\pm$ 0.039) \\
    Scattering Networks \cite{scattering}    & 0.43 (single model)       \\
    Maxout Networks \cite{maxou}        & 0.45 (single model)       \\
    Stochastic Pooling \cite{stoch}     & 0.47 (single model)       \\     
    \hline
  \end{tabular}
  \end{center}  
  \caption{\label{tab1}Results on unmodified MNIST dataset}  
\end{table}

We obtained the results of the first experiment using a simple MATLAB implementation, mainly because the GPU implementation does not allow for any arbitrary choice for the number of maps per layer. However, for the second task of MNIST discussed below and the rest of experiments we used the GPU implementation. In addition, for the first task of MNIST we used quadratic error. For the rest of the experiments, however, the cross-entropy error was minimized.

For the second task, we used data translation, rotation and scaling. Image translation was achieved by cropping the original images to 24$\times$24 
patches picked at random locations. Random rotations of maximum 15 degrees and random scaling of maximum 15\% of the original size were also 
used during training. For this task, we opted for a network with two larger convolutional layers. For the first layer, 5$\times$5 filters were used to produce 64 maps, 
and 4$\times$4 filters were used in the second one to produce 128 maps. LDNNs with 4 groups and 4 discriminants per group perform the classification part. We did not use any 
preprocessing 
for this task. We trained 5 networks for 100, 15 and 15 epochs with corresponding learning rates of 1e-3, 1e-4 and 1e-5. An error rate of 0.19\% was achieved by voting the 
results of these 5 networks. Table \ref{tab2} compares this result to other methods. The result of DropConnect \cite{dropc} (0.21\%) is obtained after 1000 
epochs of training, which is nearly 10 times more than our training epochs. Our best results over different runs of the algorithm was 0.22\% for a single model network. For comparison, the 0.23\% error rate reported in \cite{mc} obtained by voting of 35 single networks. 
\begin{table}[h]
  \begin{center}
  \begin{tabular}{ | c | c | }
  \hline
                   & Error rates \\
    Method         & Voting (\%) ( Mean (\%) $\pm$ std. dev)\\ \hline
    Proposed structure       & 0.19 (0.27 $\pm$ 0.036)  \\
    DropConnect \cite{dropc} & 0.21 (0.28 $\pm$ 0.032)  \\
    Dropout \cite{dropc} & 0.27 (0.28 $\pm$ 0.016) \\
    MC-DNN \cite{mc}  & 0.23 (0.44 $\pm$ 0.060) \\
    \hline    
  \end{tabular}
  \end{center}  
  \caption{\label{tab2}Results on augmented MNIST dataset}  
\end{table}
\subsection{CIFAR10}
CIFAR10 \cite{cifar} is a collection of 60000 tiny 32$\times$32 images of 10 categories (50000 for training and 10000 for test). Our setup for this dataset 
consists of 2 convolutional layers followed by two locally connected layers. There are 64 maps in each convolutional layer and 32 maps in each 
locally connected layer. Filters are 5$\times$5 in convolutional layers and 3$\times$3 in locally connected layers (the same as {\tt 'layers-conv-local-13pct.cfg'} of \cite{cudac}). LDNNs with 7 groups and 7 discriminants per group were used for classification. Training data was also augmented with image 
translations, which is done by taking 24$\times$24 cropped versions of the original images at random locations. A common preprocessing for this dataset
is to subtract the per pixel mean of the training set from every image \cite{HintonDropout}. We trained 5 networks for 500, 20 and 20 epochs with corresponding learning 
rates of 1e-3, 1e-4 and 1e-5 for convolutional layers and 1e-2, 1e-3 and 1e-4 for LDNN layer. We obtained 9.39\% error rate by voting these 5 single 
networks. The state-of-the-art result for this task is 9.32\% reported by Wan {\em et al.} \cite{dropc} and obtained by voting of 12 networks. They also used the same GPU implementation 
of \cite{cudac} to obtain this number. Their model, however, is based on {\tt 'layers-conv-local-11pct.cfg'} setting, which is a slightly more complex model. This setting contains two 
extra response normalization layers and the first locally connected layer contains 64 maps (vs. 32 in our setting). Another notable difference is that they trained their network for over 1000 epochs. In comparison, our 
network converged in 540 epochs, which is nearly half of their number of epochs. It must be noted that our method achieves better average error rate compared to other models that use multiple networks. The result of Maxout networks \cite{maxou} obtained by a much bigger network which has 3 convolutional layers with 192, 384 and 384 maps respectively. The classification layer also has 2500 nodes (500 maxout hidden nodes with 5 linear units for each of them). They perform global contrast normalization and ZCA whitening before training their model.
\begin{table}[h]  
  \begin{center}
  \begin{tabular}{ | c | c | }
  \hline
                 & Error rates: \\
    Method         & Voting (\%) ( Average (\%) $\pm$ std. dev)\\ \hline
    Proposed structure          & 9.39 (10.56 $\pm$ 0.18) \\
    DropConnect \cite{dropc} (5 nets)   & 9.41 (11.10 $\pm$ 0.13) \\
    DropConnect \cite{dropc} (12 nets)   & 9.32 (voting error) \\    
    Dropout \cite{dropc}        & 9.83 (11.52 $\pm$ 0.18)  \\
    Maxout Networks \cite{maxou}  & 9.38 (single model)\\
    MC-DNN \cite{mc} & 11.21 (17.42 $\pm$ 1.96) \\
    \hline
  \end{tabular}
  \end{center}
  \caption{\label{tab3}Results on CIFAR10 dataset}
\end{table}
\subsection{NORB}
NORB \cite{lecun5} is a collection of stereo images in 6 classes. The training set contains 10 folds of 29160 images. It is common practice to use only first two folds for training. The test set contains 2 folds totalizing 58320. The original images are 108$\times$108. However, we scaled them down to 
48$\times$48 similar to \cite{mc}.
The layer configuration for this task is similar to CIFAR10. LDNNs also have 7 groups and 7 discriminants per group. We trained this structure for 75 and 20 epochs with learning 
rates of 1e-3 and 1e-4. Data translation, rotation and scaling were also used during training. Image translation was obtained by randomly cropping
the training images to 44$\times$44. We trained 4 networks and used voting for final classification. We obtained 3.09\% error rate on this task. The state-of-the-art for this task is 3.03\% reported by Wan {\em et al.} \cite{dropc} as shown in Table \ref{tab4}. They did not use image translation 
as they found that it hurts the performance. In addition, their model is slightly more complex and requires more training epochs (150). Here again, our method achieves the best average error rate.
\begin{table}[h]
  \begin{center}
  \begin{tabular}{| c | c |}
  \hline
                 & Error rates: \\
    Method         & Voting (\%) ( Average (\%) $\pm$ std. dev)\\ \hline
    Proposed structure            & 3.09 (3.60 $\pm$ 0.13) \\
    DropConnect \cite{dropc} & 3.23 (4.14 $\pm$ 0.06) \\
    Dropout \cite{dropc}    & 3.03 (3.96 $\pm$ 0.16) \\
    Multi-column DNN \cite{mc}  & 3.57 (4.72 $\pm$ 0.16) \\
    \hline
  \end{tabular}
  \end{center}  
  \caption{\label{tab4}Results on NORB dataset}  
\end{table}
\subsection{SVHN}
SVHN \cite{svhn} is another digit classification task similar to MNIST. This dataset contains 604388 images for training and validation. The test set contains 
26032 images, which are RGB images of size 32 $\times$ 32. Generally, SVHN is a more difficult task than MNIST because of the large variations in the images. 
It is common to apply local contrast normalization to each 3 RGB channel of the input image in order to reduce the effect of variations of the images \cite{svhn2}. We did not perform any kind of preprocessing for this dataset. We simply 
converted the color images to grayscale by removing hue and saturation information. The feature extractor is similar to CIFAR10. We used 
locally connected layers with 64 maps for this case. LDNNs with 4 groups and 4 discriminants per group perform the classification. We did a careful annealing in this case and trained 
the model for 200, 20, 20 and 10 epochs with learning rates of 1e-3, 1e-4, 1e-5 and 5e-6. Image translation, rotation and scaling were also applied during 
training. Images were cropped to 28$\times$28 at random locations for image translation. For the last 10 epochs, we turned off image rotation and scaling. Four networks were trained using this setting and we obtained a 1.92\% error rate after voting.
Table \ref{tab5} compares this number to other results. The 1.94\% reported by Wan {\em et al.} \cite{dropc} is obtained after 150 epochs. This faster convergence is probably because of their 
preprocessing scheme. However, our model still achieves the best average error rate.
\begin{table}[h]
  \begin{center}
  \begin{tabular}{ | c | c | }
  \hline
                 & Error rates: \\
    Method         & Voting (\%) ( Average (\%) $\pm$ std. dev)\\ \hline
    Proposed structure           & 1.92 (2.14 $\pm$ 0.037) \\
    DropConnect \cite{dropc}     & 1.94 (2.23 $\pm$ 0.039) \\
    Dropout \cite{dropc}         & 1.96 (2.25 $\pm$ 0.034) \\
    \hline
  \end{tabular}  
  \end{center}
  \caption{\label{tab5}Results on SVHN dataset}  
\end{table}
\subsection{CIFAR100}
CIFAR100 \cite{cifar} is similar to CIFAR10, but it contains tiny images of 100 classes. We used the same setup as CIFAR10. The only difference
is that we used LDNNs with 4 groups and 4 discriminants per group instead of 7 and 7. Per pixel mean subtraction and image translation were also applied. Four networks were 
trained and an error rate of 36.17\% was obtained after voting, which shows that our structure is able to handle datasets with many classes.

\section{Conclusion}
\label{sec:conclusion}
We believe that the LDNN network architecture and training can become a favorable alternative to MLPs for supervised learning with artificial neural networks. The LDNN classifier has several advantages over MLPs: First, LDNNs allow for a simple and intuitive initialization of the network weights before supervised learning that avoids the herd-effect. Second, due to the single adaptive layer, learning can use larger step-sizes in gradient descent. We demonstrated empirically that LDNNs are significantly faster to train and are more accurate than MLPs. Similar to MLPs, the LDNN classifier also requires the choice of model complexity. The number of conjunctions (number of positive training clusters) and the number of logistic sigmoid functions per conjunction (number of negative training clusters) need to be chosen. However, the complexity of the model could be chosen automatically by either using a validation set, as commonly done for SVM training, or by initializing the LDNN in different ways. For instance, sequential covering algorithms can be used to generate a set of rules~\cite{Mitchell}. Each rule is a conjunction and the final classification is a disjunction of these conjunctions which can easily be converted to a LDNN classifier and fine tuned using gradient descent. 

While LDNNs are similar in architecture to modular neural networks~\cite{Lu1999}, they are significantly more accurate owing to the unified training of the network that we introduced. LDNNs outperformed RFs in 13 of the 16 datasets and outperformed SVMs  in 12 of the 16 datasets. Based on these results and observations, we believe that LDNNs should be considered as a state-of-the art classifier that provides a viable alternative to RFs and SVMs.  Further improvements in accuracy can be possible by using cross-entropy instead of the square error criterion or by using adaptive step sizes for training LDNNs. Another possibility is to use more powerful nonlinear discriminants such as conic sections in (\ref{eq:sigmoid}). 

Finally, our deep structure is a novel combination of a powerful automatic feature learner and LDNN as an efficient classifier. The whole structure is jointly optimized using back-propagation via the chain rule. We demonstrated its reliability through different experiments on MNIST, CIFAR10, NORB, SVHN and CIFAR100 datasets. We showed that it is possible to achieve state-of-the-art or near state-of-the-art results on these datasets using the proposed structure. Furthermore, in most of the cases, the average error rate of our structure is better than state-of-the-art methods.

\bibliographystyle{IEEEtran}
\bibliography{ldnnbib}

%

%
%
%
%



\end{document}